\documentclass{article} 
\usepackage{iclr2016_conference,times}
\usepackage{hyperref}
\usepackage{amsmath}
\usepackage{amsfonts}
\usepackage{url}

\usepackage{amssymb}
\usepackage{subfig}
\usepackage{xcolor}
\usepackage{lipsum}
\usepackage[demo]{graphicx}
\usepackage[neverdecrease]{paralist}
\usepackage{mathtools}
\usepackage{tikz}
\usetikzlibrary{arrows}

\newsavebox{\measurebox}

\title{Reasoning in Vector Space:\\An Exploratory Study of Question Answering}

\author{
Moontae Lee\thanks{This research was conducted while the first author held a summer internship in Microsoft Research, Redmond, and the last author was a Visiting Researcher there.} \\
Department of Computer Science\\
Cornell University University\\
Ithaca, NY 14850, USA \\
\texttt{moontae@cs.cornell.edu} \\
\And
Xiaodong He, Wen-tau Yih, Jianfeng Gao \& Li Deng \\
Microsoft Research\\
Redmond, WA 98052, USA \\
\texttt{\{xiaohe,scottyih,jfgao,deng\}@microsoft.com} \\
\AND
Paul Smolensky$^*$ \\
Department of Cognitive Science\\
Johns Hopkins University\\
Baltimore, MD 21218, USA\\
\texttt{smolensky@jhu.edu}
}

\iclrfinalcopy 

\usepackage{enumitem}
\begin{document}

\maketitle

\begin{abstract}
Question answering tasks have shown remarkable progress with distributed vector representation. 
In this paper, we investigate the recently proposed Facebook bAbI tasks which consist of twenty different categories of questions that require complex reasoning.
Because the previous work on bAbI are all end-to-end models, errors could come from either an imperfect understanding of semantics or in certain steps of the reasoning.
For clearer analysis, we propose two vector space models inspired by Tensor Product Representation (TPR) to perform knowledge encoding and logical reasoning based on common-sense inference.
They together achieve near-perfect accuracy on all categories including positional reasoning and path finding that have proved difficult for most of the previous approaches.
We hypothesize that the difficulties in these categories are due to the multi-relations in contrast to uni-relational characteristic of other categories. 
Our exploration sheds light on designing more sophisticated dataset and moving one step toward integrating transparent and interpretable formalism of TPR into existing learning paradigms.

%
\end{abstract}

\section{Introduction}
Ideal machine learning systems should be capable not only of learning rules automatically from training data, but also of transparently incorporating existing principles.
While an end-to-end framework is suitable for learning without human intervention, existing human knowledge is often valuable in leveraging data toward better generalization to novel input.
Question answering (QA) is one of the ultimate tasks in Natural Language Processing (NLP) on which synergy between the two capabilities could enable better understanding and reasoning.

Recently the Facebook bAbI tasks were introduced to evaluate complex reading comprehension via QA~(\cite{Weston2015}); these have received considerable attention.
Understanding natural questions, for example in WebQuestions tasks~(\cite{ berant-EtAl:2013:EMNLP}), requires significant comprehension of the semantics, yet reasoning out the answers is then relatively simple~(e.g.,~\cite{ bordes-chopra-weston:2014:EMNLP2014, Yih2015}).
In contrast, the synthetic questions in bAbI require rather complex reasoning over multiple computational steps while demanding only minimal semantic understanding.
As the previous work on bAbI consists only of end-to-end models~(\cite{Weston2014, Kumar2015, Sukhbaatar2015, Peng2015}), it is unclear whether incorrect answers arise from an imperfect semantic understanding, inadequate knowledge encoding, or insufficient model capacity~(\cite{Dupoux2015}).
This is partly because the current paradigms based on neural networks have no interpretable intermediate representations which modelers can use to assess the knowledge present in the vectorial encoding of the system's understanding of the input sentences.
Our approach, in contrast, can illuminate what knowledge is caputred in each representation via the formalism of TPR.

Tensor Product Representation (TPR), proposed by \cite{Smolensky1990, Smolensky2006}, is a mathematical method to represent complex structures from basic vectorial building blocks, so called \textit{fillers} and \textit{roles}. 
For example, one can encode a binary tree by \textit{binding} filler vectors corresponding to the left- and right-child entities to role vectors corresponding to the `left child' and `right child' positions, respectively. 
Arbitrary trees can be represented by recursively applying the same method. 
As an outer product (i.e., tensor product) realizes the binding operation, both filler and role components are decodable from the resulting representation via the inner product; this is called \textit{unbinding}. 
TPR is known to be capable of various applications such as tree operations, grammar processing and lambda-calculus evaluation (\cite{Smolensky2012}).

In this paper, we endeavor to disentangle the problem cleanly into semantic parsing, knowledge encoding, and logical reasoning.
Proposing two vector-space models inspired by TPR, we first provide an in-depth analysis of the bAbI dataset by clustering, based solely on their logical properties, the twenty question categories defined by bAbI.
Such analysis enables us to conjecture why most existing models, in spite of their complexity, have failed to achieve good accuracy on \textit{positional reasoning} and \textit{path finding} tasks, whereas \cite{Peng2015} achieved successful results.
If the bAbI tasks turn out to be considerably simpler than intended for its ultimate purpose of providing a major step towards ``AI-complete question answering'', then more elaborated tasks will be required to test the power of proposed QA models such as memory networks.

As a further contribution, we also develop the foundation of a theory that maps inference for logical reasoning to computation over TPRs, generalizing our models under the rigorous TPR formalism. Due to the page limit, this theoretical foundation is relegated to the supplementary materials (\cite{Smolensky2016}). The experimental results show that accurate inference based on common-sense knowledge is transparently attainable in this formalism. We hope our exploration can contribute to the further improvement of end-to-end models toward the transparency and interpretability. To the best of our knowledge, our in-depth analysis of bAbI and of logical reasoning over distributed vectorial representations are each the first of their kind.


\section{Related Work}
Since the seminal work of \cite{Bengio2003}, researchers have paid increasing attention to various distributed representations in continuous vector spaces.
In the computer science literature, Skip-gram/CBoW (\cite{Mikolov2013}) and GloVe (\cite{Pennington2014}) are popular models that are trained based on the distributional similarities in word co-occurrence patterns;  they have been frequently utilized as initial embeddings for a variety of other NLP tasks.
In the cognitive science literature, on the other hand, BEAGLE (\cite{Jones2007}) and DVRS (\cite{Ustun2014}) are trained differently, with random initializations and circular convolution.
They assign two vectors for each word: an environmental vector to describe physical properties and a lexical vector to indicate meaning.

Whereas such representations are known to provide a useful way to incorporate prior linguistic knowledge, their usefulness is not clear for reasoning-oriented tasks.
In other contexts, \cite{Grefenstette2013} shows how to simulate predicate logic with matrices and tensors.
Similarly, \cite{Rocktaschel2014} try to find low-dimensional embeddings which can model first-order logic in a vectorial manner.
These models are only concentrated on general logic problems without considering NLP tasks.
Note that vectorial encodings are necessary in many machine learning models such as neural networks.
Reasoning based on linguistic cues in vector space uniquely characterizes our paper among these relevant work.

The tasks in bAbI have been studied mainly within the context of the Memory Network (MemNN) model, which consists of an array of representations called ``memory'' and four learnable modules: the I-module encodes the input into feature representation, the G-module updates relevant memory slots, the O-module performs inferences to compute output features given the input representation and the current memory, and finally the R-module decodes the output feature-based representation to the final response.
Since the proposal of the basic MemNN (\cite{Weston2014}) model, the Adaptive/Nonlinear MemNN (\cite{Weston2015}), DMN (\cite{Kumar2015}), and MemN2N (\cite{Sukhbaatar2015}) models have been developed by varying certain parts of these modules.
Nonetheless, none of these models except \cite{Peng2015} successfully accomplish either positional reasoning or path finding tasks.
Our speculation about the performance by \cite{Peng2015} will be given in a later section based on our bAbI analysis.

\section{Models and Analysis}
The bAbI dataset consists of twenty different types of questions where each question category is claimed to be atomic and independent from the others (\cite{Weston2015}). 
In this section, we investigate clusters of categories with sample QA problems, analyzing what kinds of logical properties are shared across various types.
We also elucidate, based on our vector space models, why it is difficult to achieve good accuracy on certain categories: positional reasoning and path finding.

\subsection{Containee-Container Relationship}

\paragraph{Supporting Facts (1, 2, 3)}
The first three question categories of bAbI ask for the current or previous locations of actors and objects based on the statements given prior to the question.
Category 1--3 questions respectively require precisely one, two, or three supporting facts to reason out the proper answers.
Figure \ref{fig:Sample1} illustrates  sample statements and questions extracted from real examples in the training set.
Reasoning in Category 1 implicitly requires a simple common-sense reasoning rule that \textit{``An actor cannot exist in two different locations at the same time.''}
In order to answer the questions in Category 2, we implicitly need another rule that \textit{``An object that belongs to an actor follows its owner's location.''}
Further, if an item is dropped at one particular location, it will permanently stay in that location until someone grabs it and moves around with it later.

\begin{figure}
\small
\centering
\sbox{\measurebox}{%
\begin{minipage}[b][\ht\measurebox][s]{.395\textwidth}
\centering
\vspace{-3px}
    \fcolorbox{black}{white}{%
    \minipage[t]{\dimexpr0.9\linewidth-2\fboxsep-2\fboxrule\relax}
\subfloat
{\bf \underline{Category 1: Single Supporting Fact}}\\
01: Mary moved to the bathroom.\\
02: John went to the hallway.\\
{\color{blue}03: Where is Mary?} 	{\color{red}bathroom}	{\color{green}1}\\    
04: Daniel went back to the hallway.\\
05: Sandra moved to the garden.\\
{\color{blue}06: Where is Daniel?} 	{\color{red}hallway}	{\color{green}4}    
    \endminipage}\hfill
\vfill
\vspace{3px}
    \fcolorbox{black}{white}{%
    \minipage[t]{\dimexpr0.9\linewidth-2\fboxsep-2\fboxrule\relax}
\subfloat
{\bf \underline{Category 2: Two Supporting Facts}}\\
01: Mary went to the kitchen.\\
02: Sandra journeyed to the office.\\
03: Mary got the football there.\\
04: Mary travelled to the garden.\\
{\color{blue}05: Where is the football?} 	{\color{red}garden}	{\color{green}3 4}\\
06: John travelled to the office.\\
07: Sandra moved to the garden.\\
{\color{blue}08: Where is the football?} 	{\color{red}garden}	{\color{green}3 4}\\
09: Mary dropped the football.\\
10: Mary journeyed to the kitchen.\\
{\color{blue}11: Where is the football?} 	{\color{red}garden}	{\color{green}9 4}
    \endminipage}\hfill
\end{minipage}}
\usebox{\measurebox}
    \fcolorbox{black}{white}{%
    \minipage[t]{\dimexpr0.595\linewidth-2\fboxsep-2\fboxrule\relax}
    \subfloat
{\bf \underline{Category 3: Three Supporting Facts}}\\    
01: Sandra went back to the hallway.\\
02: Daniel took the apple.\\
03: John travelled to the kitchen.\\
04: Daniel travelled to the bedroom.\\
05: Daniel got the football there.\\
06: Daniel went to the hallway.\\
{\color{blue}07: Where was the apple before the hallway?} 	{\color{red}bedroom}	{\color{green}2 6 4}\\
08: Mary went back to the bedroom.\\
09: Daniel discarded the football.\\
10: Daniel got the football.\\
11: Mary went to the garden.\\
12: Daniel travelled to the office.\\
13: Daniel went back to the bedroom.\\
{\color{blue}14: Where was the football before the bedroom?} 	{\color{red}office}	{\color{green}10 13 12}\\
15: Daniel went back to the hallway.\\
16: Mary went back to the bathroom.\\
17: Daniel dropped the apple.\\
18: Sandra journeyed to the kitchen.\\
{\color{blue}19: Where was the apple before the office?} 	{\color{red}hallway}	{\color{green}17 12 6}
  \endminipage}
\caption{Sample statements(black), questions(blue), answers(red), and clues(green) for Categories 1, 2, and 3.}
\label{fig:Sample1}
\end{figure}

While two independent relations, \textit{pick/drop} and \textit{move}, seem to be involved in parallel in the Category 2 tasks, these questions can be all uniformly answered under the transitivity of \textit{a containee belongs to a container}.
If an actor moves to a location, he/she (a containee) now belongs to that location (a container).
Similarly, if an actor acquires an object, the item (a containee) newly belongs to that actor (a container).
Transitivity then logically implies that the object belongs to the location occupied by the owner.

\begin{table}[h]
\begin{center}
\small
\setlength\tabcolsep{3.5pt}
  \begin{tabular}{ c l l l l }
    \hline
    \textbf{\#} & \textbf{Statements/{\color{blue}Questions}} & \textbf{Relational Translations/{\color{red}Answers}} & \multicolumn{2}{l}{\textbf{Encodings/{\color{green}Clues}}}  \\
    \hline
    1 & Mary went to the kitchen. & \textit{Mary belongs to the kitchen (from nowhere).} & $mk^T$ & $m(k \circ n)^T$ \\
    3 & Mary got the football there. &  \textit{The football belongs to Mary.} & $fm^T$ & $fm^T $\\
    4 & Mary travelled to the garden. & \textit{Mary belongs to the garden (from the kitchen).}  & $mg^T$ & $m(g \circ k)^T$\\
    5 & {\color{blue}Where is the football?} & {\color{red}garden} & {\color{green}3, 4} & \\
    9 & Mary dropped the football. & \textit{The football belongs to where Mary belongs to.} & $fg^T$ & $fg^T$\\
    10 & Mary journeyed to the kitchen. & \textit{Mary belongs to the kitchen (from the garden).} & $mk^T$ & $m(k \circ g)^T$\\
    11 & {\color{blue}Where is the football?} & {\color{red}garden} & {\color{green}9, 4} &\\
    \hline
  \end{tabular}
\end{center}
\vspace{-5px}
\caption{Sample \textit{containee-belongs\_to-container} translations and corresponding encodings about Mary from Category 2. Symbols in encodings are all $d$-dimensional vectors for actors ($m$ary), objects ($f$ootball), and locations($n$owhere, $k$itchen, $g$arden). Translations and encodings for Category 3 are also specified with the parentheses and circle operation, respectively.}
\label{tbl:Sample1}
\end{table}
Knowing that every actor and object is unique without any ambiguity, one can encode such containee-conatainer relationships by the following model using distributed representations.
Assume all entities: actors, objects, and locations are represented by $d$-dimensional unit vectors in $\mathbb{R}^d$.\footnote{Topologically speaking, the unit hypersphere can be constructed by adding one more point (``at infinity'') to Euclidean space. Thus sampling from the hypersphere does not limit the generality of representations.}
Then each statement is encoded by a second-order tensor (or matrix) in which the containee vector is bound to the container vector via the fundamental binding operation of TPR, the tensor (or outer) product\footnote{In TPR terms, the containee corresponds to a filler, and the container corresponds to a role.} --- in tensor notation,
$(\textit{containee})\otimes(\textit{container})$, or in matrix notation, 
$(\textit{containee}) (\textit{container})^{T}$
--- and then stored in a slot in a memory.
When an item is dropped, we perform an inference to store the appropriate knowledge in memory.
For the example in Table \ref{tbl:Sample1}, the container of the football at Statement 9 --- the garden --- is determined after figuring out the most recent owner of the football, Mary; transitivity is implemented through simple matrix multiplication of the encodings of Statement 3 (locating the football) and  Statement 4 (locating the football's current owner, Mary): 
\[
(fm^T) \cdot (mg^T) = f(m^T \cdot m)g^T = fg^T \qquad (\because m^T m =  ||m||_2^2 = 1)
\]

Finally, Category 3 asks the trajectory of items considering the previous locations of actors.
Thus the overall task is to understand the relocation sequence of each actor and from this to reconstruct the trajectory of item locations.
Whereas MemNNs introduced an additional vector for each statement for encoding a time stamp, we define another binding operation $\circ : \mathbb{R}^d \times \mathbb{R}^d \longrightarrow \mathbb{R}^d$.
This binding operation maps a pair of ($n$ext, $p$rev) location vectors into a $d$-dimensional vector via a $d \times 2d$ temporal encoding matrix $U$ like the following:
\[
n \circ p = U
\begin{bmatrix}
n\\
p
\end{bmatrix}
\in \mathbb{R}^d.
\]

In Table \ref{tbl:Sample1}, the second expression in the Encodings column specifies temporal encodings that identify location transitions: Statement 4, translated as \textit{Mary belongs to the garden (from the kitchen)}, is encoded as $m(g \circ k)^T$. 
We can now reason to the proper answers for the questions in Figure \ref{fig:Sample1} by the following inference steps, using basic encodings (for C1 \& C2) and temporal encodings (for C3):
\setdefaultleftmargin{1cm}{0.2cm}{}{}{}{}
\begin{enumerate}
  \item[C1.] Where is Mary?
    \begin{enumerate} 
      \item[(a)] Left-multiply by $m^{T}$ all statements prior to time 3. (Yields $m \cdot m^{T} b^{T}$, $m^{T} \cdot j h^{T}$.)
      \item[(b)] Pick the most recent container where 2-norms of the multiplications in (a) are approximately 1.0. (Yields $b^{T}$; $m^{T} j$ is small.)
      \item[(c)] Answer by finding the location corresponding to the result representation. $\Rightarrow$ \textbf{\textit{bathroom}}
    \end{enumerate}

  \item[C2.] Where is the football?
    \begin{enumerate} 
      \item[(a)] Left-multiply by $f^{T}$ all statements prior to the current time. (Yields $f^{T} \cdot m k^{T}$, $f^{T} \cdot s o^{T}$, $f^{T} \cdot f m^{T}$, $f^{T} \cdot m g^{T}$.)
      \item[(b)] Pick the most recent container where 2-norms of the multiplications in (a) are approximately 1.0. (Yields $m^{T}$.)
      \item[(c)] If the container is an actor (e.g., Mary in statement 3),
        \begin{itemize}
          \item Find the most recent container of the actor by left-multiplying by $m^{T}$ (Yields $g^{T}$.)
          \item Answer by the most recent container. $\Rightarrow$ \textbf{\textit{garden}} for the questions at time 5 and 8.
        \end{itemize}
      \item[(d)] If the container is a location (e.g., garden in statement 9), simply answer by the container.        
    \end{enumerate}

  \item[C3.] Where was the apple before the hallway?
    \begin{enumerate} 
      \item[(a)] Left-multiply by $a^{T}$ all existing temporal encodings prior to time 7. (Yields $a^{T} \cdot s (h \ \circ \ n)^{T}$, $a^{T} \cdot ad^{T}$, ... .)
      \item[(b)] Pick the earliest container (the start of the trajectory). $\Rightarrow$ \textit{Daniel} in statement 2.
      \item[(c)] Find the containers of Daniel by left-multiplying by $d^{T}$ the temporal encodings between time 2 and 7. (Yields $d^{T} \cdot ad^{T}$, $d^{T} \cdot j (k \circ n)^{T}$, $d^{T} \cdot d (b \circ n)^{T}$, $d^{T} \cdot fd^{T}$, $d^{T} \cdot d(h \circ b)$, ... .)
      \item[(d)] By multiplying by the pseudo-inverse $U^{\dagger}$, unbind $2d$-dimensional vectors between time 4 and 7. (Yields $U^{\dagger}(b \circ n) \approx [b; n]$, then $[h; b]$.)
      \item[(e)] Reconstruct the item trajectory in sequence. $\Rightarrow \textit{nowhere} \rightarrow \textit{bedroom} \rightarrow \textit{hallway}$
      \item[(f)] Answer with (the most recent) location which is prior to the hallway. $\Rightarrow$ \textbf{\textit{bedroom}}
    \end{enumerate}  
\end{enumerate}

\paragraph{Three Argument Relations (5)}
In this category, there is a new type of statement which specifies ownership transfer: an actor gives an object to another actor. 
Since now some relations involve three arguments, (\textit{source-actor}, \textit{object}, \textit{target-actor}), we need to encode an ownership trajectory instead of a location trajectory. 

\begin{table}[h]
\begin{center}
\small
\setlength\tabcolsep{3.5pt}
  \begin{tabular}{ c l l l}
    \hline
    \textbf{\#} & \textbf{Statements/{\color{blue}Questions}} & \textbf{Relational Translations/{\color{red}Answers}} & \textbf{Encodings/{\color{green}Clues}}  \\
    \hline
    1 & Jeff took the milk there. & \textit{The milk belongs to Jeff (from None).} & $m(j * n)^T$\\
    2 & Jeff gave the milk to Bill. & \textit{The milk belongs to Bill (from Jeff).} & $m(b * j)^T$\\
    3 & {\color{blue}Who did Jeff give the milk to?} & {\color{red}Bill}	& {\color{green}2}\\
    4 & Daniel travelled to the office. & \textit{Daniel belongs to the office.} & $do^T$\\
    5 & Daniel journeyed to the hallway. & \textit{Daniel belongs to the hallway.} & $dh^T$\\
    6 & {\color{blue}Who received the milk?} & {\color{red}Bill}	& {\color{green}2}\\
    7 & Bill went to the kitchen. & \textit{Bill belongs to the kitchen.} & $bk^T$\\
    8 & Fred grabbed the apple there. & \textit{The apple belongs to Fred (from none).} & $a(f * n)^T$\\
    9 & {\color{blue}What did Jeff give to Bill?} & {\color{red}milk}	& {\color{green}2}\\
    \hline
  \end{tabular}
\end{center}
\vspace{-5px}
\caption{Sample \textit{containee-belongs\_to-container} translations and corresponding encodings for an example from Category 5. Symbols in encodings are all $d$-dimensional vectors for actors ($n$obody, $j$eff, $d$aniel, $b$ill, $f$red), objects ($m$ilk, $a$pple), and locations ($o$ffice, $k$itchen).\protect\footnotemark}
\label{tbl:Sample2}
\end{table}
\footnotetext{To avoid notational confusion, we modify the name of an actor (from Mary to Daniel) and a location (from the bathroom to the office) from the real example in Category 5.}

Analogously to the $\circ$ operation used for Category 3, we realize the $*$ operation by defining a map $* : \mathbb{R}^d \times \mathbb{R}^d \longrightarrow \mathbb{R}^d$.
This new binding operation maps a pair of ($n$ext, $p$rev) owner vectors into a $d$-dimensional vector via a $d \times 2d$ matrix $V$ in the exactly same fashion: $n * p = V[n; p] \in \mathbb{R}^d$. 
Due to the similarity in encoding, the inference is also analogous to the inference for Category 3.

\begin{enumerate}
  \item[C5.] Three questions of Table \ref{tbl:Sample2}?
    \begin{enumerate}
       \item[(a)] Find the owners of the milk by left-multiplying by $m^{T}$ the encodings prior to time 3. 
       \item[(b)] Unbind the owner transitions by multiplying them by the pseudo-inverse $V^{\dagger}$.
       \item[(c)] Reconstruct the ownership trajectory for the milk. $\Rightarrow \textit{Nobody} \rightarrow \textit{Jeff} \rightarrow \textit{Bill}$
       \item[(d)] Answer accordingly each question based on the trajectory.
    \end{enumerate}
 \end{enumerate}

\begin{figure}
\small
\centering
  \sbox{\measurebox}{%
    \begin{minipage}[b][\ht\measurebox][s]{.435\textwidth}
    \centering
      \fcolorbox{black}{white}{%
      \minipage[t]{\dimexpr0.95\linewidth-2\fboxsep-2\fboxrule\relax}
\subfloat
{\bf \underline{Category 6: Yes/No Questions}}\\
01: Daniel went back to the hallway.\\
02: John got the apple there.\\
{\color{blue}03: Is Daniel in the hallway?} 	{\color{red}yes}	{\color{green}1}\\    
04: John dropped the apple.\\
05: Mary got the apple there.\\
{\color{blue}06: Is Daniel in the hallway?} 	{\color{red}yes}	{\color{green}1}\\    
07: Daniel moved to the bedroom.\\
08: Sandra travelled to the hallway.\\
{\color{blue}09: Is Daniel in the hallway?} 	{\color{red}no}	{\color{green}7}
    \endminipage}\hfill
    \vfill
    \vspace{3px}
    \fcolorbox{black}{white}{%
    \minipage[t]{\dimexpr0.95\linewidth-2\fboxsep-2\fboxrule\relax}
\subfloat
{\bf \underline{Category 8: List/Sets}}\\
01: Mary took the milk there.\\
02: Mary went to the office.\\
{\color{blue}03: What is Mary carrying?} 	{\color{red}milk}	{\color{green}1}\\
04: Mary took the apple there.\\
05: Sandra journeyed to the bedroom.\\
{\color{blue}06: What is Mary carrying?} 	{\color{red}milk,apple}	{\color{green}1 4}\\
07: Mary put down the milk.\\
08: Mary discarded the apple.\\
{\color{blue}09: What is Mary carrying?} 	{\color{red}nothing}	{\color{green}1 7 4 8}
    \endminipage}\hfill
    \end{minipage}}
   \usebox{\measurebox}
    \begin{minipage}[b][\ht\measurebox][s]{.555\textwidth}
    \centering
       \fcolorbox{black}{white}{%
       \minipage[t]{\dimexpr0.95\linewidth-2\fboxsep-2\fboxrule\relax}
\subfloat
{\bf \underline{Category 7: Counting}}\\    
01: Mary took the apple there.\\
02: John travelled to the office.\\
{\color{blue}03: How many objects is Mary carrying?} 	{\color{red}one}	{\color{green}1}\\
04: Mary travelled to the bathroom.\\
05: Sandra went back to the bedroom.\\
{\color{blue}06: How many objects is Mary carrying?} 	{\color{red}one}	{\color{green}1}\\
07: Mary got the football there.\\
08: Mary went to the office.\\
{\color{blue}09: How many objects is Mary carrying?} 	{\color{red}two}	{\color{green}1 7}\\
10: Mary passed the apple to John.\\
11: Mary left the football.\\
{\color{blue}12: How many objects is Mary carrying?} 	{\color{red}none}	{\color{green}1 7 10 11}
    \endminipage}\hfill
    \vfill
    \vspace{3px}
    \fcolorbox{black}{white}{%
    \minipage[t]{\dimexpr0.95\linewidth-2\fboxsep-2\fboxrule\relax}
\subfloat
{\bf \underline{Category 9: Simple Negation}}\\
01: Sandra travelled to the garden.\\
02: Sandra is no longer in the garden.\\
{\color{blue}03: Is Sandra in the garden?} 	{\color{red}no}	{\color{green}2}\\
04: Sandra is in the garden.\\
05: Sandra journeyed to the hallway.\\
{\color{blue}06: Is Sandra in the hallway?} 	{\color{red}yes}	{\color{green}5}
    \endminipage}\hfill
    \end{minipage}
\vspace{110px}
\caption{Sample statements(black), questions(blue), answers(red), and clues(green) for Category 6, 7, 8, and 9. Answer types are different from the previous categories.}
\label{fig:Sample2}
\end{figure}

Though no more complex examples or distinct categories exist in the dataset, it is clear that our encoding scheme is capable of inferring the full trajectory of item location considering both relocation of actors and transfers of ownership.
In such cases, both $\circ$ and $*$ will be used at the same time in encoding. (e.g., encoding for time 5 will be then $d(h \ \circ \ o)^T$.
Note also that there may be multiple transfers between the same pair of actors in a history prior to the given question. 
While any of them could be appropriate evidence to justify different answers, the ground-truth answers in the training set  turned out to be all based on the most recent clues.

\paragraph{Answer Variations (6, 7, 8, 9)}
As shown in Figure \ref{fig:Sample2}, the responses to questions of Categories 6-9 require different measures of the inferred element.
For example, the statements in Category 6 are structurally equivalent to the statements in Category 2, while the questions concern only a current location, similar to Category 1.
However, each question is formulated in a binary yes/no format, confirming \textit{``Is Daniel in the hallway?''} instead of asking \textit{``Where is Daniel?"}.
Category 7 is isomorphic to Category 5 in the sense that actors can pick up, drop, and pass objects to other actors.
However, each question inquires the number of objects currently belonging to the given actor.
On the other hand, a response in Category 8 must give the actual names of objects instead of counting their number. 
The statements in this category are based not on Category 5, but on Category 2 which is simpler due to the lack of ownership transfer.
Lastly, statements in Category 9 can contain a negative quantifier such as \textit{`no'} or \textit{`no longer'}.
Responses confirm or not the location of actors via yes/no dichotomy as for Category 6.
However, the overall story is based on the simplest Category 1.

Since answer measures are the only differences of these categories from Category 1, 2, 3, and 5, no additional encodings or inferences are necessary.
However, there are several caveats in formulating actual answers: 1) For yes/no questions, we should know the answers must be either \textit{yes} or \textit{no} in advance based on the training examples. 2) When counting the number of belongings, the answer must use English number words rather than Arabic numerals. 
3) When enumerating the names of belongings, names must be sequenced by their order of acquisition.  
4) A negative quantifier is realized by binding the initial default location \textit{nowhere} back to the given actor. Note that there is no double negation.

\paragraph{Statement Variations (10, 11, 12, 13)}
Statements in Categories 10-13 contain more challenging  linguistic elements such as conjunctions (\textit{and/or}) or pronouns (\textit{he/she/they}). 
While statements in Category 10 is structurally similar to Category 1's, an actor can be located in \textit{either} one \textit{or} another location. 
Due to such uncertainty, some questions must be answered indefinitely by \textit{`maybe'}.
On the other hand, each statement in Category 12 can contain multiple actors conjoined by \textit{`and'} to indicate that these actors all carry out the action.
Aside from such conjunctions, statements and questions are isomorphic to Category 1's.
Statements in Categories 11/13 can consist of a singular/plural pronoun to indicate single/multiple actors mentioned earlier.
Since coreference resolution is itself a difficult problem, all pronouns are limited to refer only to actors mentioned in the immediately prior statement.

To encode conjunctions, we can still leverage the same method: conjoin two objects by another bilinear binding operation $\star: \mathbb{R}^d \times \mathbb{R}^d \longrightarrow \mathbb{R}^d$, and unbind similarly via the pseudo-inverse of the corresponding matrix.
In our implementation, every statement is encoded using such a binding operation.
For instance, the first two statements of the given Category 10 example are encoded into $j(k \star k)^T$ and $b(s \star o)^T$, with $\star$ encoding \textit{or}. 
If two locations unbound from the target actor are identical, we output a yes/no definite answer, whereas two different locations imply the indefinite answer \textit{`maybe'} if one of the unbound locations matches the queried location.
For the conjunction \textit{and} in Category 12,  exactly the same formalism is applicable for conjoining actors instead.
Whereas a singular pronoun appearing at time $t$ in Category 11 is simply replaced by the actor mentioned at time $t-1$, we also use $\star$-binding to provide the multiple coreference needed for Category 13.
For instance, the first statement in the given Category 13 example is encoded as $(m \star d)b^T$ and the same encoding is substituted for \textit{`they'} to represent the actors in the following statement.

\begin{figure}
\small
\centering
  \sbox{\measurebox}{%
    \begin{minipage}[b][\ht\measurebox][s]{.52\textwidth}
    \centering
      \fcolorbox{black}{white}{%
      \minipage[t]{\dimexpr0.95\linewidth-2\fboxsep-2\fboxrule\relax}
\subfloat
{\bf \underline{Category 10: Indefinite Knowledge}}\\
01: Julie travelled to the kitchen.\\
02: Bill is either in the school or the office.\\
{\color{blue}03: Is Bill in the office?} 	{\color{red}maybe}	{\color{green}2}\\    
04: Bill went back to the bedroom.\\
05: Bill travelled to the kitchen.\\
{\color{blue}06: Is Bill in the kitchen?} 	{\color{red}yes}	{\color{green}5}
    \endminipage}\hfill
    \vfill
    \vspace{3px}
    \fcolorbox{black}{white}{%
    \minipage[t]{\dimexpr0.95\linewidth-2\fboxsep-2\fboxrule\relax}
\subfloat
{\bf \underline{Category 12: Conjunction}}\\
01: Daniel and Sandra went back to the kitchen.\\
02: Daniel and John went back to the hallway.\\
{\color{blue}03: Where is Daniel?} 	{\color{red}hallway}	{\color{green}2}\\
04: Daniel and John moved to the bathroom.\\
05: Sandra and Mary travelled to the office.\\
{\color{blue}06: Where is Daniel?} 	{\color{red}bathroom}	{\color{green}4}
    \endminipage}\hfill
    \end{minipage}}
   \usebox{\measurebox}
    \begin{minipage}[b][\ht\measurebox][s]{.465\textwidth}
    \centering
       \fcolorbox{black}{white}{%
       \minipage[t]{\dimexpr0.95\linewidth-2\fboxsep-2\fboxrule\relax}
\subfloat
{\bf \underline{Category 11: Basic Coreference}}\\    
01: Mary went back to the bathroom.\\
02: After that she went to the bedroom.\\
{\color{blue}03: Where is Mary?} 	{\color{red}bedroom}	{\color{green}1 2}\\
04: Daniel moved to the office.\\
05: Afterwards he moved to the hallway.\\
{\color{blue}06: Where is Daniel?} 	{\color{red}hallway}	{\color{green}4 5}
    \endminipage}\hfill
    \vfill
    \vspace{3px}
    \fcolorbox{black}{white}{%
    \minipage[t]{\dimexpr0.95\linewidth-2\fboxsep-2\fboxrule\relax}
\subfloat
{\bf \underline{Category 13: Compound Coreference}}\\
01: Mary and Daniel went to the bathroom.\\
02: Then they journeyed to the hallway.\\
{\color{blue}03: Where is Daniel?} 	{\color{red}hallway}	{\color{green}1 2}\\
04: Sandra and John moved to the kitchen.\\
05: Then they moved to the hallway.\\
{\color{blue}06: Where is John?} 	{\color{red}hallway}	{\color{green}4 5}
    \endminipage}\hfill
    \end{minipage}
\vspace{80px}
\caption{Sample statements(black), questions(blue), answers(red), and clues(green) for Category 10, 11, 12, and 13. Statement types are different from the previous categories.}
\label{fig:Sample3}
\vspace{-10px}
\end{figure}

\paragraph{Deduction/Induction (15, 16, 18, 20)}
While the statements and questions in these categories seem different at first glance, their goals are all to reason using  a transitivity-like rule.
Categories 15 creates a food chain among various animals, and Category 18 yields a partial/total order of sizes among various objects.
Whereas inference in these two categories is deductive, Categories 16 and 20 require inductive inference.
In all four categories, every statement is easily represented by a containee-container relation obeying transitivity. 
For instance, the Category 15 example of Figure \ref{fig:Sample4} is encoded by $\{mc^T, wm^T, cs^T, sw^T\}$.
Then the answer for the first question: \textit{``What is Jessica afraid of?''} will be answered by left-multiplying these by the transpose of $j=m$ and finding the one whose norm is approximately 1.0, which is $mc^T$.
Thus the result $j^T \cdot (mc^T) = m^T (mc^T) = (m^T m)c^T = c^T$ produces the desired answer \textit{cat}.
Similarly, in Category 18, if question encoding (e.g., \textit{``Does the chocolate fit in the box?''} = $cb^T$) is achievable by some inner products of statement encodings, the answer must be \textit{`yes'}, otherwise, \textit{`no'}.


On the other hand, in Category 16, transitivity is applied reversely as a \textit{container-containee} fashion.
For instance, ``Lily is a $\ell$ion" is encoded by $\ell l^T$, whereas ``Lily is green'' is encoded by $lg^T$.
In encoding ``x is-a Y'', we put the more general concept at the left side of the outer-product binding $Y x^{T}$; to encode ``x has-property Z'' we use $x Z^{T}$. 
This allows us to induce a property for the general category Y based on the single observation of one of its members, via simple matrix multiplication, just as transitive inference was implemented above: 
$(\ell l^T)  \cdot (lg^T) = \ell g^T$, meaning \textit{``$\ell$ion is green.''}
Similarly in Category 20, there exists precisely one statement which describes a property of an actor (e.g., \textit{``Sumit is bored.''} = $bs^T$).
Then a statement describes the actor's relocation (e.g., \textit{``Sumit journeyed to the garden.''} = $sg^T$), yielding an inductive conclusion by matrix multiplication: \textit{``Being boring makes people go to the garden.''} = $(bs^T) \cdot (sg^T) = bg^T$.
The inductive reasoning also generalizes to other actions (e.g., the reason for later activity, \textit{``Sumit grabbed the football.''} = $sf^T$, is also being bored, because $(bs^T) \cdot (sf^T) = bf^T$).

\begin{figure}
\small
\centering
  \sbox{\measurebox}{%
    \begin{minipage}[b][\ht\measurebox][s]{.45\textwidth}
    \centering
      \fcolorbox{black}{white}{%
      \minipage[t]{\dimexpr0.95\linewidth-2\fboxsep-2\fboxrule\relax}
\subfloat
{\bf \underline{Category 15. Basic Deduction}}\\
01: Mice are afraid of cats.\\
02: Emily is a mouse.\\
03: Wolves are afraid of mice.\\
04: Cats are afraid of sheep.\\
05: Winona is a cat.\\
06: Sheep are afraid of wolves.\\
07: Jessica is a mouse.\\
08: Gertrude is a sheep.\\
{\color{blue}09: What is Jessica afraid of?} 	{\color{red}cat}	{\color{green}7 1}\\
{\color{blue}10: What is Emily afraid of?} 	{\color{red}cat}	{\color{green}2 1}\\
{\color{blue}11: What is Jessica afraid of?} 	{\color{red}cat}	{\color{green}7 1}\\
{\color{blue}12: What is Winona afraid of?} 	{\color{red}sheep}	{\color{green}5 4}
    \endminipage}\hfill
    \vfill
    \vspace{3px}
    \fcolorbox{black}{white}{%
    \minipage[t]{\dimexpr0.95\linewidth-2\fboxsep-2\fboxrule\relax}
\subfloat
{\bf \underline{Category 16: Basic Induction}}\\
01: Bernhard is a lion.\\
02: Julius is a lion.\\
03: Lily is a lion.\\
04: Bernhard is green.\\
05: Lily is green.\\
06: Brian is a lion.\\
07: Greg is a swan.\\
08: Greg is gray.\\
09: Julius is yellow.\\
{\color{blue}10: What color is Brian?} 	{\color{red}green}	{\color{green}6 3 5}
    \endminipage}\hfill
    \end{minipage}}
   \usebox{\measurebox}
    \begin{minipage}[b][\ht\measurebox][s]{.535\textwidth}
    \centering
       \fcolorbox{black}{white}{%
       \minipage[t]{\dimexpr0.95\linewidth-2\fboxsep-2\fboxrule\relax}
\subfloat
{\bf \underline{Category 18: Reasoning about Size}}\\    
01: The suitcase is bigger than the container.\\
02: The container fits inside the box.\\
03: The chest is bigger than the chocolate.\\
04: The suitcase fits inside the box.\\
05: The chest fits inside the box.\\
{\color{blue}06: Does the chocolate fit in the box?} 	{\color{red}yes}	{\color{green}5 3}\\
{\color{blue}07: Does the chocolate fit in the box?} 	{\color{red}yes}	{\color{green}5 3}\\
{\color{blue}08: Does the box fit in the container?} 	{\color{red}no}	{\color{green}1 4}\\
{\color{blue}09: Is the box bigger than the chocolate?} 	{\color{red}yes}	{\color{green}5 3}\\
{\color{blue}10: Does the box fit in the chocolate?} 	{\color{red}no}	{\color{green}3 5}
    \endminipage}\hfill
    \vfill
    \vspace{4px}
    \fcolorbox{black}{white}{%
    \minipage[t]{\dimexpr0.95\linewidth-2\fboxsep-2\fboxrule\relax}
\subfloat
{\bf \underline{Category 20: Reasoning about Motivations}}\\
01: Sumit is bored.\\
{\color{blue}02: Where will Sumit go?} 	{\color{red}garden}	{\color{green}1}\\
03: Yann is hungry.\\
{\color{blue}04: Where will Yann go?} 	{\color{red}kitchen}	{\color{green}3}\\
05: Yann went back to the kitchen.\\
{\color{blue}06: Why did Yann go to the kitchen?} 	{\color{red}hungry}	{\color{green}3}\\
07: Sumit journeyed to the garden.\\
{\color{blue}08: Why did Sumit go to the garden?} 	{\color{red}bored}	{\color{green}1}\\
09: Yann picked up the apple there.\\
{\color{blue}10: Why did Yann get the apple?} 	{\color{red}hungry}	{\color{green}3}\\
11: Sumit grabbed the football there.\\
{\color{blue}12: Why did Sumit get the football?} 	{\color{red}bored}	{\color{green}1}
    \endminipage}\hfill
    \end{minipage}
\vspace{125px}
\caption{Sample statements(black), questions(blue), answers(red), and clues(green) for Category 15, 16, 18, and 20. Categories 15 and 18 create chains from smaller/weaker to stronger/larger, whereas Categories 16 and 20 from general ones to specific ones.}
\label{fig:Sample4}
\end{figure}
\paragraph{Prior Knowledge (4, 14)}
Though statements in Category 4 looks quite dissimilar from those in the other categories, they can be eventually modeled by a uni-relational reasoning chain based on the containee-container relation, provided we know that \textit{`north'} and \textit{`south'} are opposite to each other. 
Thus the first two statements in the first Category 4 example in Figure \ref{fig:Sample5} yield $\{ko^T, gk^T\}$, from which we infer $(gk^T) \cdot (ko^T) = go^T$ \textit{``The office is north of the garden.''}
While the questions are all simple knowledge confirmation, note that a relational word (e.g., \textit{`east'}) might never appear in the prior statements, as illustrated in the second example of Category 4 in Figure \ref{fig:Sample5}. 
However the most important point is that two non-collinear relations (e.g., \textit{`north', `east'}) never appear together in the same example.

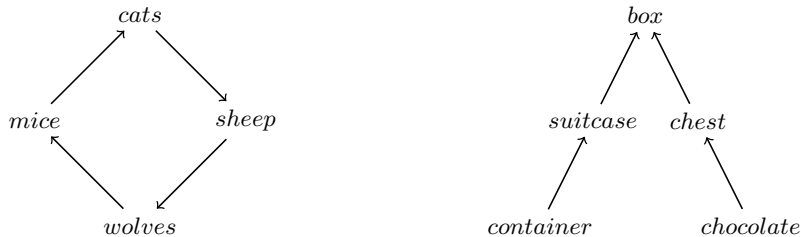
\begin{figure}
\small
\centering
\hspace{30px}
\begin{minipage}{.45\textwidth}
\begin{tikzpicture}[scale=.7]
  \node (one) at (0:2cm) {$sheep$};
  \node (a) at (90:2cm) {$cats$};
  \node (b) at (180:2cm) {$mice$};
  \node (zero) at (270:2cm) {$wolves$};
  \draw [semithick,<-] (one) -- (a);
  \draw [semithick,<-] (a) -- (b);
  \draw [semithick, <-] (b) -- (zero);
  \draw [semithick, <-] (zero) -- (one);
\end{tikzpicture}
\end{minipage}
\begin{minipage}{.45\textwidth}
\begin{tikzpicture}[scale=.7]
  \node (a) at (0,2) {$box$};
  \node (b) at (-1,0) {$suitcase$};
  \node (c) at (1,0) {$chest$};
  \node (d) at (-2,-2) {$container$};
  \node (e) at (2,-2) {$chocolate$};
  \draw [semithick,->] (e) -- (c);
  \draw [semithick,->] (d) -- (b);
  \draw [semithick, ->] (c) -- (a);
  \draw [semithick, ->] (b) -- (a);
\end{tikzpicture}
\end{minipage}
\caption{The circular food chain (Category 16) and the partial order (Category 18) corresponding to the examples in Figure \ref{fig:Sample4}. The arrows imply \textit{afraid-of} and \textit{fits-inside} relations, respectively.}
\end{figure}

\begin{figure}
\small
\centering
  \sbox{\measurebox}{%
    \begin{minipage}[b][\ht\measurebox][s]{.50\textwidth}
    \centering
      \fcolorbox{black}{white}{%
      \minipage[t]{\dimexpr0.95\linewidth-2\fboxsep-2\fboxrule\relax}
\subfloat
{\bf \underline{Category 4. Two Argument Relation}}\\
01: The office is north of the kitchen.\\
02: The garden is south of the kitchen.\\
{\color{blue}03: What is north of the kitchen?} 	{\color{red}office}	{\color{green}1}\\
------------------------------------------------------------\\
01: The kitchen is west of the garden.\\
02: The hallway is west of the kitchen.\\
{\color{blue}03: What is the garden east of?} 	{\color{red}kitchen}	{\color{green}1}
    \endminipage}\hfill
    \vfill
    \vspace{13px}
    \fcolorbox{black}{white}{%
    \minipage[t]{\dimexpr0.95\linewidth-2\fboxsep-2\fboxrule\relax}
\subfloat
{\bf \underline{Category 14: Time Manipulation}}\\    
01: Yesterday Julie went back to the park.\\
02: Julie went to the bedroom this morning.\\
03: Bill journeyed to the cinema yesterday.\\
04: This morning Bill went back to the park.\\
{\color{blue}05: Where was Bill before the park?} 	{\color{red}cinema}	{\color{green}4 3}\\
06: This evening Julie went to the school.\\
07: This afternoon Julie went back to the park.\\
{\color{blue}08: Where was Julie before the bedroom?} 	{\color{red}park}	{\color{green}2 1}
    \endminipage}\hfill
    \end{minipage}}
   \usebox{\measurebox}
    \begin{minipage}[b][\ht\measurebox][s]{.49\textwidth}
    \centering
       \fcolorbox{black}{white}{%
       \minipage[t]{\dimexpr0.95\linewidth-2\fboxsep-2\fboxrule\relax}
\subfloat
{\bf \underline{Category 17: Positional Reasoning}}\\    
01: The triangle is above the pink rectangle.\\
02: The blue square is to the left of the triangle.\\
{\color{blue}03: Is the pink rectangle to the right of the blue square?} 	{\color{red}yes}	{\color{green}1 2}\\
-----------------------------------------------------------\\
01: The red sphere is below the yellow square.\\
02: The red sphere is above the blue square.\\
{\color{blue}03: Is the blue square below the yellow square?} 	{\color{red}yes}	{\color{green}2 1}
    \endminipage}\hfill
    \vfill
    \vspace{4px}
    \fcolorbox{black}{white}{%
    \minipage[t]{\dimexpr0.95\linewidth-2\fboxsep-2\fboxrule\relax}
\subfloat
{\bf \underline{Category 19: Path Finding}}\\
01: The bedroom is south of the hallway.\\
02: The bathroom is east of the office.\\
03: The kitchen is west of the garden.\\
04: The garden is south of the office.\\
05: The office is south of the bedroom.\\
{\color{blue}06: How do you go from the garden to the bedroom?} 	{\color{red}n,n}	{\color{green}4 5}
    \endminipage}\hfill
    \end{minipage}
\vspace{100px}
\caption{Sample statements(black), questions(blue), answers(red), and clues(green) for Category 4, 14, 17, and 19. Categories 4 and 17 contains two different examples separated by a horizontal line.}
\label{fig:Sample5}
\end{figure}

On the other hand, statements in Category 14 are no longer chronologically ordered.
In order to infer a correct locational trajectory without repeating statements multiple times, we predefine four vectors for each time stamp: $y$esterday, this $m$orning, this $a$fternoon, and this $e$vening, and bind location with the corresponding stamp instead of the previous location. 
For example, the encoding for the statement at time 2 now becomes $j(b \circ m)^T$ instead of $j(b \circ p)^T$.
Knowing the correct order of these four time stamps, which could be learned from the training examples, we can easily reorder by unbinding time stamps.


\subsection{Multiple Relationships}
\paragraph{Path Finding (19)}
Our goal in this category is to find the path from one location to another location in a Manhattan-grid-like sense. 
Note that if $A$ is north of $B$, and $B$ is north of $C$, then the right path from $A$ to $C$ in grid must be \textit{`\textbf{n}orth, \textbf{n}orth'} rather than simply \textit{'\textbf{n}orth'}. 
We assume given four $d \times d$ non-singular matrices $N, E, W, S$ encoding four different directions satisfying $N=S^{-1}$ and $E=W^{-1}$. 
Then

\begin{table}[h]
\begin{center}
\small
\setlength\tabcolsep{3.5pt}
  \begin{tabular}{ c l l l c }
    \hline
    \textbf{\#} & \textbf{Statements/{\color{blue}Questions}} & \textbf{Translations/{\color{red}Answers}/{\color{green}Clues}} & \textbf{Encodings} & \textbf{Seq} \\
    \hline
    1 & The bedroom is south of the hallway. & \textit{Decides $b$ given the initial $h$.} & $b = Sh$ & (1)\\
    2 & The $\beta$athroom is east of the office. & \textit{Defer until we know either $o$ or $\beta$.} & $\beta = Eo$ & (3)\\
    3 & The kitchen is west of the garden. & \textit{Defer until we know either $g$ or $k$.} & $k = Wg$ & (5)\\
    4 & The garden is south of the office. & \textit{Defer until we know either $o$ or $g$.} & $g = So$ & (4)\\
    5 & The office is south of the bedroom. & \textit{Decides $o$ given $b$.} & $o = Sb$ & (2)\\
    6 & {\color{blue}How do you go from the garden to the bedroom?} & {\color{red}n,n} \quad {\color{green}4, 5} & $b=Xg$ & (6)\\
    \hline
  \end{tabular}
\end{center}
\vspace{-5px}
\caption{Sample multi-relational translations and corresponding encodings from Category 19. Symbols in encodings are either $d$-dimensional object vectors ($h$allway, $b$edroom, $o$ffice, $\beta$athroom, $g$arden, $k$itchen) or $d \times d$ directional matrices ($S$outh, $E$ast, $W$est, $N$orth). The last column shows the sequence of actual running order.}
\label{tbl:Sample5}
\end{table}

After initializing the first object in the right-hand side (e.g., \textit{`hallway'}) by a random vector, we decide the rest of the object vectors in sequence by multiplying the directional matrix (or its inverse in case that the right-hand side is unknown and the left-hand side is known). 
In case that both sides are unknown, we defer such a statement by putting it into a queue. 
In fact, the solution path $X$ can be determined either by selecting, of all combinations of two directions $\{\textit{NN, NE, NW, NS, ... SN, SE, SW, SS}\}$, the one which best satisfies $b=Xg$ (in the example of Table \ref{tbl:Sample5}) or by solving this equation based on iterative substitutions.
Note also that we need to know that ($n$, $e$, $w$, $s$) in the answers correspond to (north, east, west, south), respectively, which could be learned from training data.

\paragraph{Positional Reasoning (17)}
While this category could be seen similar to Path Finding, each question only asks a relative position between two objects. 
For instance, if \textit{``$r$ is below $s$''}, and \textit{``$b$ is below $r$''}, then the position of $b$ with respect to $s$ must be simply \textit{`below'} rather than \textit{`below, below'}. 
Even if an object is mentioned to be left of another object, it could be also located in left-above or left-below of another object. 
Due to these subtleties, we here adopt redundant representations with four $d \times d$ singular matrices $(A, B, L, R)$ corresponding to four directions: (above, below, left, right).
For this directional subsumption, in contrast to the non-singularity of the directional matrices for Category 19, we now strictly enforce idempotency to these matrices (i.e., $A^n =...= A^2 = A \neq I$). 
Then we define the following four $4d \times 4d$ block matrices and encode each statement  with these matrices in the same manner as for Category 19.
\[
\mathcal{A} = 
\begin{bmatrix}
    A & 0 & 0 & 0 \\
    0 & I & 0 & 0 \\
    0 & 0 & I & 0\\
    0 & 0 & 0 & I\\
\end{bmatrix}
\;\;
\mathcal{B} = 
\begin{bmatrix}
    I & 0 & 0 & 0 \\
    0 & B & 0 & 0 \\
    0 & 0 & I & 0\\
    0 & 0 & 0 & I\\
\end{bmatrix}
\;\;
\mathcal{L} = 
\begin{bmatrix}
    I & 0 & 0 & 0 \\
    0 & I & 0 & 0 \\
    0 & 0 & L & 0\\
    0 & 0 & 0 & I\\
\end{bmatrix}
\;\;
\mathcal{R} = 
\begin{bmatrix}
    I & 0 & 0 & 0 \\
    0 & I & 0 & 0 \\
    0 & 0 & I & 0\\
    0 & 0 & 0 & R\\
\end{bmatrix}
\]

In this encoding, each of the four $d$-dimensional subspaces of $\mathbb{R}^{4d}$ plays a role of indicating relative positions with respect to (above, below, left, right), independently. 
Carrying out the encoding of \textit{``$r$ is below $s$''}, $r = \mathcal{B}s$, ensures that the components of $r$ and $s$  differ only in the dimensions from $(d+1)$ to $2d$ (from the $B$ block of $\mathcal{B}$); that is, $r_{k} = s_{k}$ for $k = 1, 3, 4$ (where $s_i$ indicates the $i$-th $d$-dimensional sub-block of $s$).
This is actually inconsistent with the encoding of \textit{``$s$ is above $r$''}, which demands that $s$ and $r$ differ only in their first sub-block. 
Thus in order to determine whether or not $s$ is indeed above $r$, it is necessary to check whether $r_2 = B s_2$ as well as whether $s_1 = A r_1$.
If either condition is satisfied, we can confirm \textit{`$s$ is above to $r$'} . 
Similarly, horizontal relations must be checked on both the third and fourth $d$-dimensional sub-blocks.

\section{Experimental Results}
We implement our models and algorithms under the analysis given in the previous section. 
Due to the small vocabulary (mostly less than or equal to four elements among actors, objects, locations, and actions) and non-ambiguous grammars, a simple dependency parser\footnote{We use Stanford Dependency Parser. \url{http://nlp.stanford.edu/software/stanford-dependencies.shtml}} and basic named entity recognition enable us to achieve 100\% accurate semantic parsing.
Then we translate every statement into a representation based on the appropriate containee-container or multiway relation, and then store it in an array of memory slots.
The logical reasoning after semantic parsing and knowledge representation no longer refers to the original text symbols.
\begin{table}[h]
\begin{center}
\small
  \begin{tabular}{ c c c c c c c c c c c }
    \hline
        \textbf{Type} & \textbf{C1} & \textbf{C2} & \textbf{C3} & \textbf{C4} & \textbf{C5} & \textbf{C6} & \textbf{C7} & \textbf{C8} & \textbf{C9} & \textbf{C10}\\
    \hline
    Accuracy & 100\% & 100\% & 100\% & 100\% & 99.3\% & 100\% & 96.9\% & 96.5\% & 100\% & 99\%\\
    Model & MNN & MNN & MNN & MNN & DMN & MNN & DMN & DMN & DMN & SSVM\\
    \hline
    \textbf{Type} & \textbf{C11} & \textbf{C12} & \textbf{C13} & \textbf{C14} & \textbf{C15} & \textbf{C16} & \textbf{C17} & \textbf{C18} & \textbf{C19} & \textbf{C20}\\
    \hline
    Accuracy & 100\% & 100\% & 100\% & 100\% & 100\% & 100\% & 72\% & 95\% & 36\% & 100\% \\
    Model & MNN & MNN & MNN & DMN & MNN & MNN & Multitask & MNN & MNN & MNN\\
    \hline
  \end{tabular}
\end{center}
\caption{Best accuracies for each category and the model that achieved the best accuracy. MNN indicates Strongly-Supervised MemNN trained with the clue numbers, and DMN indicates Dynamic MemNN, and SSVM indicates Structured SVM with the coreference resolution and SRL features. Multitask indicates multitask training.}
\label{tbl:Result1}
\end{table}

In contrast to all previous models reported in Table \ref{tbl:Result1}, in Table \ref{tbl:Result2} we also report test accuracy on the training data to measure how well our models incorporate common sense.
Note that testing on the training data is available because our training procedure only parses the appropriate semantic components such as actors, objects, locations, actions, and the forms of answers without using given answers and clues for tuning the model parameters.
\begin{table}[h]
\begin{center}
\small
  \begin{tabular}{ c c c c c c c c c c c }
    \hline
        \textbf{Type} & \textbf{C1} & \textbf{C2} & \textbf{C3} & \textbf{C4} & \textbf{C5} & \textbf{C6} & \textbf{C7} & \textbf{C8} & \textbf{C9} & \textbf{C10}\\
    \hline
    Training & 100\% & 100\% & 100\% & 100\% & 99.8\% & 100\% & 100\% & 100\% & 100\% & 100\%\\
    Test & 100\% & 100\% & 100\% & 100\% & 99.8\% & 100\% & 100\% & 100\% & 100\% & 100\%\\
    \hline
    \textbf{Type} & \textbf{C11} & \textbf{C12} & \textbf{C13} & \textbf{C14} & \textbf{C15} & \textbf{C16} & \textbf{C17} & \textbf{C18} & \textbf{C19} & \textbf{C20}\\
    \hline
    Training & 100\% & 100\% & 100\% & 100\% & 100\% & 99.4\% & 100\% & 100\% & 100\% & 100\%\\
    Test & 100\% & 100\% & 100\% & 100\% & 100\% & 99.5\% & 100\% & 100\% & 100\% & 100\%\\
    \hline
  \end{tabular}
\end{center}
\caption{Accuracies on training and test data on our models. We achieve near-perfect accuracy in almost every category including positional reasoning and path finding.}
\label{tbl:Result2}
\end{table}

Note that the imperfect accuracy in Category 16 is due to the ambiguity of evidence. 
As given in Figure \ref{fig:Sample4}, one can answer the color of \textit{Brian} as \textit{`yellow'} because the latest evidence tells \textit{Julius who is a lion is yellow.}
Similarly, in Category 5, the 8th story consists of incorrect/inconsistent answers at time 14 and 17 (for training), as they ignore the most recent ownership transfers and choose some old history as ground-truth answers. (The 63rd and 186th stories in the test data also consist of incorrect answers, at time 27 and 22, respectively)
Other than these two categories, we achieve perfect accuracies performing common-sense operations only on representations in memory.

As the experimental results show, there is a clear distinction between two sets of tasks. 
Tasks in most categories can be modeled by a containee-container-like relationship respecting a   transitivity-like inference rule, whose goals are to create a linear/circular chain. 
On the other hand, positional reasoning and path finding require multiple relationships where each corresponding pair (e.g., north vs. south) has its own transitivity structure, operating independently of other pairs (e.g. east vs. west). 
We hypothesize that this difference poses a major difficulty for most of Memory Network models to perform an accurate inference for positional reasoning and path finding.

Recently, Neural Reasoner (NR) by \cite{Peng2015} improves the accuracy for these two difficult categories by a large margin, achieving 97.9\% and 87.0\% when using 10k training set. \footnote{5 All accuracy values of various models reported in the experimental section of the present paper are based on a 1k training set. Neural Reasoner achieves 66.4\% and 17.3\% when using the 1k dataset.}
Different from other memory network models, NR has multiple reasoning layers. 
Starting from the initial statements and questions, NR constructs new statements and questions at the next layer, and repeats this process recursively over multiple layers. 
As both positional reasoning and path finding require generating inferences from, and new versions of, relevant statements for each relationship (e.g., ``x is north of '' can become ``y is south of x''), the abilities to generate new facts and to derive final answers by integrating them from multiple relationships could be a key reason why NR is successful, like our TPR-based reasoner. 
While NR in experiment is simplified so that all new facts maintain the same initial representations, the question representation changes for each layer considering all existing facts and the previously evolved question.
Due to the simplicity of the task, we conjecture that evolving representations of the question could be sufficient to comprise the key ingredient for each multi-relationship.
However, it seems that training such multiple layers requires a large amount of training data, yielding drastically different performance of NR on two different dataset sizes.

\section{Conclusion}
The major contributions of this paper are two-fold. 
First, we throughly analyze the recently acclaimed bAbI question-answering tasks by grouping the twenty categories based on their relational properties.
Our analysis reveals that most categories except positional reasoning and path finding are governed by uni-relational characteristics.
As these turn out to support inference in a similar manner under  transitivity, it could be dangerous to evaluate the capacity of network models based only on their performance on bAbI.
In contrast, two more difficult categories require the capability of performing multi-relational reasoning, a capability which is apparently missing in most previous models.
One could later develop a more sophisticated dataset that needs substantially harder reasoning by introducing multiple relationships.
Second, we propose two vector space models which can perform logistic reasoning for QA with distributed representations.
While TPR has been used for various problems such as tree/grammar encoding and lambda-calculus evaluation, logical reasoning is a new area of application that requires iterative processing of TPRs.
In subsequent work, we will generalize the vector-space approach for multi-relational problems.
We hope these studies shed light on the viability of developing further reasoning models which can perform inference with existing knowledge in an interpretable and transparent manner. 


%

\newpage
\nocite{Weston2014}
\nocite{Weston+Bordes2015}
\nocite{Kumar2015}
\nocite{Sukhbaatar2015} 
\nocite{Dupoux2015}
\small
\bibliography{iclr2016_conference}
\bibliographystyle{iclr2016_conference}

\end{document}